%% file: main.tex
\documentclass{article}

\usepackage{microtype}
\usepackage{graphicx}
\usepackage{subfigure}
\usepackage{booktabs} % for professional tables
\usepackage{xcolor,soul}
\usepackage{hyperref}
\usepackage{enumitem}
\usepackage{multirow}
\usepackage{array}

\usepackage[accepted]{icml2023}

\usepackage{amsmath}
\usepackage{amssymb}
\usepackage{mathtools}
\usepackage{amsthm}

\usepackage[capitalize,noabbrev]{cleveref}

\theoremstyle{plain}

\theoremstyle{definition}

\theoremstyle{remark}

\usepackage[textsize=tiny]{todonotes}

\icmltitlerunning{Plan, Eliminate, and Track}

\begin{document}

\twocolumn[
\icmltitle{Plan, Eliminate, and Track ---\\
           Language Models are Good Teachers for Embodied Agents.}



\icmlsetsymbol{equal}{*}

\begin{icmlauthorlist}
\icmlauthor{Yue Wu}{cmu}
\icmlauthor{So Yeon Min}{cmu}
\icmlauthor{Yonatan Bisk}{cmu}
\icmlauthor{Ruslan Salakhutdinov}{cmu}
\icmlauthor{Amos Azaria}{ariel}
\icmlauthor{Yuanzhi Li}{cmu,msft}
\icmlauthor{Tom M. Mitchell}{cmu}
\icmlauthor{Shrimai Prabhumoye}{nvda}
\end{icmlauthorlist}

\icmlaffiliation{cmu}{Carnegie Mellon University}
\icmlaffiliation{msft}{Microsoft Research}
\icmlaffiliation{ariel}{Ariel University}
\icmlaffiliation{nvda}{Nvidia Research}

\icmlcorrespondingauthor{Yue Wu}{ywu5@andrew.cmu.edu}

\icmlkeywords{AlfWorld, Text Games, Large Language Models, In-context prompting, Language Models, Zero-shot, Few-shot}

\vskip 0.3in
]



\printAffiliationsAndNotice{}  % leave blank if no need to mention equal contribution

\begin{abstract}
% \href{https://larel-workshop.github.io/}{Workshop Link}
Pre-trained large language models (LLMs) capture procedural knowledge about the world. 
Recent work has leveraged LLM's ability to generate abstract plans to simplify challenging control tasks, either by action scoring, or action modeling (fine-tuning). However, the transformer architecture inherits several constraints that make it difficult for the LLM to directly serve as the agent: e.g. limited input lengths, fine-tuning inefficiency, bias from pre-training, and incompatibility with non-text environments. 
% instead of knowledge transfer to a domain-specific control agent
%In addition, LM score on set of skills are often biased toward certain type of actions and therefore unreliable. For example, prompting a LM to find coffee pod and make coffee might result in the LM stuck in the loop between goto coffee pod and goto coffee machine due to intrinsic bias toward goto actions. 
% A more reliable way of generating actions is to obtain LM score on a set of admissible primitive skills, which requires lots of examples as prompts. 
% Such need for examples make the prompt length infeasible for more challenging tasks with longer observations and longer roll-outs, since current LMs have a maximum input length of $1024\sim2048$ tokens. 
To maintain compatibility with a low-level trainable actor, we propose to instead use the \textit{knowledge} in LLMs to simplify the control problem, rather than solving it. 

We propose the Plan, Eliminate, and Track (\textbf{PET}) framework. The Plan module translates a task description into a list of high-level sub-tasks. The Eliminate module masks out irrelevant objects and receptacles from the observation for the current sub-task. Finally, the Track module determines whether the agent has accomplished each sub-task.
On the AlfWorld instruction following benchmark, the \textbf{PET} framework leads to a significant 15\% improvement over SOTA for generalization to human goal specifications. %\TODO{Say something like: we obtain SOTA results?}
% Constructing agents with planning capabilities has been one of the key challenges for A.I. Tree-based planning methods have enjoyed huge success in challenging domains, such as chess and Go. However, tree-based methods require perfect simulators or accurate environment models, both of which are currently infeasible in real-world tasks. On the other hand, human can plan and execute a lot of tasks with language and common sense. In this work, we propose  We therefore try to assist the training of our policy  %We approach this challenge through planning and reasoning through language.
% \TODO{Add some numbers}
\end{abstract}

\input{sections/intro.tex}
\input{sections/related.tex}
\input{sections/method.tex}
\input{sections/results.tex}
\input{sections/analysis.tex}
\input{sections/conclusion.tex}

\nocite{langley00}

\bibliography{example_paper}
\bibliographystyle{icml2023}

%%%%%%%%%%%%%%%%%%%%%%%%%%%%%%%%%%%%%%%%%%%%%%%%%%%%%%%%%%%%%%%%%%%%%%%%%%%%%%%
%%%%%%%%%%%%%%%%%%%%%%%%%%%%%%%%%%%%%%%%%%%%%%%%%%%%%%%%%%%%%%%%%%%%%%%%%%%%%%%
% APPENDIX
%%%%%%%%%%%%%%%%%%%%%%%%%%%%%%%%%%%%%%%%%%%%%%%%%%%%%%%%%%%%%%%%%%%%%%%%%%%%%%%
%%%%%%%%%%%%%%%%%%%%%%%%%%%%%%%%%%%%%%%%%%%%%%%%%%%%%%%%%%%%%%%%%%%%%%%%%%%%%%%
\newpage
\appendix
\onecolumn
\include{sections/appendix.tex}
%%%%%%%%%%%%%%%%%%%%%%%%%%%%%%%%%%%%%%%%%%%%%%%%%%%%%%%%%%%%%%%%%%%%%%%%%%%%%%%
%%%%%%%%%%%%%%%%%%%%%%%%%%%%%%%%%%%%%%%%%%%%%%%%%%%%%%%%%%%%%%%%%%%%%%%%%%%%%%%

\end{document}

%% file: sections/intro.tex
\section{Introduction}
Humans can abstractly plan their everyday tasks without execution; for example, given the task ``Make breakfast'', we can roughly plan to first pick up a mug and make coffee, before grabbing eggs to scramble. 
Embodied agents, endowed with this capability will generalize more effectively by leveraging common-sense reasoning. %The \textbf{PET} framework in this work aims to do just that -- using pre-trained large language models (LLMs) to assist embodied agents.

Recent work \cite{deepak, inner_monologue, saycan, textgames} has used LLMs \cite{foundation} for abstract planning for embodied or gaming agents. 
These have shown incipient success in extracting procedural world knowledge from LLMs in linguistic form with posthoc alignment to executable actions in the environment. 
However, they treat LLMs as the actor, and focus on adapting LLM outputs to executable actions either through fine-tuning \cite{fewshot} or constraints \cite{saycan}. 
Using LLM as the actor works for pure-text environments with limited interactions \cite{inner_monologue, saycan} (just consisting of ``picking/placing" objects), but limits generalization to other modalities.
In addition, the scenarios considered have been largely simplified from the real world. 
\citet{saycan} provides all available objects and possible interactions at the start and limits tasks to the set of provided objects/interactions. \citet{inner_monologue} limits the environment to objects on a single table.

\label{PET:plan}
\begin{figure}[t]
    \centering
    \includegraphics[trim={0cm 0cm 0cm 0cm},width=0.45\textwidth]{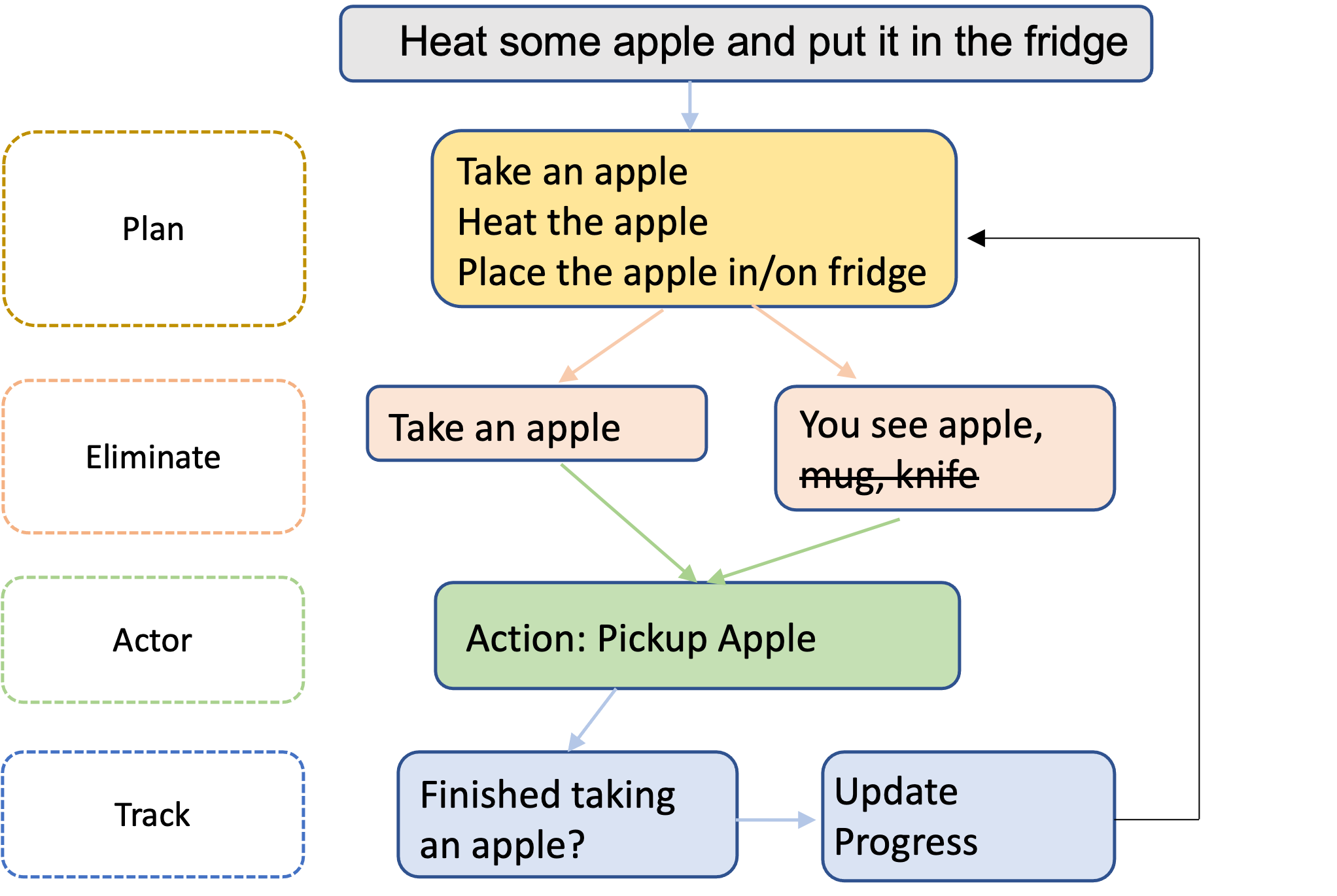} 
    \vspace{-3mm}
    \caption{\small PET framework. Plan module uses LLM to generate a high-level plan. Eliminate Module uses a QA model to mask irrelevant objects in observation. Track module uses a QA model to track the completion of sub-tasks.} % Take a apple is because the template is Take a X, not a grammatical mistake.
    \vspace{-4mm}
    \label{fig:overall}
\end{figure}

On the other hand, to successfully ``cut some lettuce'' in a real-world room, one has to ``find a knife'', which can be non-trivial since there can be multiple drawers or cabinets \cite{ogn, film, hlsm}.
A more realistic scenario leads to a diverse, complicated set of tasks or large and changing action space. Furthermore, the text description of the observation increases as a function of the number of receptacles and objects the agent sees. Combined with growing roll-outs, the state becomes too verbose to fit into any LLM. %adding to the difficulties of directly applying pre-trained language models. 

In this work, we explore alternative mechanisms to leverage the prior knowledge encoded in LLMs without impacting the trainable nature of the actor. We propose a 3-step framework (Figure~\ref{fig:overall}): Plan, Eliminate, and Track (PET). \textbf{Plan} module simplifies complex tasks by breaking them down into sub-tasks.
It uses a pre-trained LLM to generate a list of sub-tasks for an input task description employing example prompts from the training set similar to \citet{deepak,saycan}. The \textbf{Eliminate} module addresses the challenge of long observations.
It uses a zero-shot QA language model to score and mask objects and receptacles that are irrelevant to the current sub-task. The \textbf{Track} module uses a zero-shot QA language model to determine if the current sub-task is complete and moves to the next sub-task. Finally, the \textbf{Action Attention} agent uses a transformer-based architecture to accommodate for long roll-out and variable length action space. The agent observes the masked observation and takes an action conditioned on the current sub-task. 

% \TODO{I think we need to add another para either here or before this where we say that we try to mitigate the aforementioned challenges and describe it in detail like: The Elimiate module addresses the challenge of long observations and makes them shorter to fit into an LLM, Action attention helps with long rollouts by accommodating variable length, plan and track work together by first dividing the complex task into subtasks and then tracking the progress (only dividing into sub-tasks is not enough if we cannot figure out if the sub-task is completed or not).}

We focus on instruction following in indoor households on the AlfWorld \cite{alfworld} interactive text environment benchmark. Our experiments and analysis demonstrate that LLMs not only remove 40\% of task-irrelevant objects in observation through common-sense QA, but also generate high-level sub-tasks with 99\% accuracy. In addition, multiple LLMs may be used in coordination with each other to assist the agent from different aspects.

Our contributions are as follows:
\begin{enumerate}[noitemsep,nosep]
    \item \textbf{PET}: A novel framework for leveraging pre-trained LLMs with embodied agents; our work shows that each of P, E, T serves a complementary role and should be simultaneously addressed to tackle control tasks.
    \item An Action Attention agent that handles the changing action space for text environments.
    \item A 15\% improvement over SOTA for generalization to human goals via sub-task planning and tracking. 
\end{enumerate}

%% file: sections/related.tex
\section{Related Work}

\paragraph{Language Conditioned Policies}
% There is a long history of research studying how to connect language and behavior [46, 47, 48, 49, 50]. A large number of prior works have learned language-conditioned behavior via imitation learning [51, 22, 20, 13, 26, 37] or reinforcement learning [52, 53, 49, 54, 55, 56, 21, 41]. Most of these prior works focus on following low-level instructions, such as for pick-and-place tasks and other robotic manipulation primitives [20, 22, 56, 21, 13, 26], though some methods address long-horizon, compound tasks in simulated domains [57, 58, 54]. Like these latter works, we focus on completing temporally extended tasks. However, a central aspect of our work is to solve such tasks by extracting and leveraging the knowledge in large language models.
A considerable portion of prior work studies imitation learning \cite{tellex2011understanding,mei2016listen,nair2022learning,stepputtis2020language,jang2022bc,shridhar2022cliport,sharma2021skill} or reinforcement learning \cite{misra2017mapping,jiang2019language,cideron2020higher,goyal2021pixl2r,nair2022learning,akakzia2020grounding} policies conditioned on natural language instruction or goal \cite{macmahon2006walk,kollar2010toward}. While some prior research has used pre-trained language embeddings to improve generalization to new instructions \cite{nair2022learning}, they lack domain knowledge that is captured in LLMs. Our PET framework enables planning, progress tracking, and observation filtering through the use of LLMs, and is designed to be compatible with any language conditional policies above.

\paragraph{LLMs for Control}
LLMs have recently achieved success in high-level planning. \citet{deepak} shows that pre-trained LLMs can generate plausible plans for day-to-day tasks, but the generated sub-tasks cannot be directly executed in an end-to-end control environment. \citet{saycan} solves the executability issue by training an action scoring model to re-weigh LLM action choices and demonstrates success on a robot. However, LLM scores work for simple environments with actions limited to pick/place \cite{saycan}, but fails with environments with more objects and diverse actions \cite{alfworld}. \citet{song2022llm} uses GPT3 to generate step-by-step low-level commands, which are then executed by respective control policies. the work improves \citet{saycan} with more action diversity and on-the-fly re-plan. In addition, all the above LLMs require few-shot demonstrations of up to 17 examples, making the length of the prompt infeasible for AlfWorld. 
\citet{fewshot} fine-tuned a GPT2-medium model on expert trajectories in AlfWorld and demonstrated impressive evaluation results. However, LM fine-tuning requires a fully text-based environment, consistent expert trajectories, and a fully text-based action space. Such requirements greatly limit the generalization to other domains, and even to other forms of task specification. We show that our PET framework achieves better generalization to human goal specifications which the agents were not trained on.

\paragraph{Hierarchical Planning with Natural Language}
Due to the structured nature of natural language, \citet{andreas2017modular} explored associating each task description to a modular sub-policy. Later works extend the above approach by using a single conditional policy \cite{mei2016listen}, or by matching sub-tasks to templates \cite{oh2017zero}. Recent works have shown that LLMs are proficient high-level planners \cite{deepak,saycan,lin2022grounded}, and therefore motivates us to revisit the idea of hierarchical task planning with progress tracking. To our knowledge, PET is the first work combining a zero-shot subtask-level LLM planner and zero-shot LLM progress tracker with a low-level conditional sub-task policy.

\paragraph{Text Games}
Text-based games are complex, interactive simulations where the game state and action space are in natural lanugage. They are fertile ground for language-focused machine learning research. In addition to language understanding, successful play requires skills like memory and planning, exploration (trial and error), and common sense. The AlfWorld \cite{alfworld} simulator extends a common text-based game simulator, TextWorld \citet{cote2018textworld}, to create text-based analogs of each ALFRED scene. 

\paragraph{Agents for Large Action Space}
\citet{he2015deep} learns representation for state and actions with two different models and computes the Q function as the inner product of the representations. While this could generalize to large action space, they only considered a small number of actions.

\citet{fulda2017can,saycan} explore action elimination in the setting of affordances. \citet{zahavy2018learn} trains a model to eliminate invalid actions on Zork from external environment signals. However, the functionality depends on the existence of external elimination signal.

%% file: sections/method.tex
\section{Plan, Eliminate, and Track}
%\vspace{-1mm}
\label{sec:PET}

In this section, we explain our 3-step framework: Plan, Eliminate, and Track (PET). 
In \textbf{Plan} module ($\mathcal{M}_{\textbf{P}}$), a pre-trained LLM generates a list of sub-tasks for an input task description using samples from the training set as in-context examples. 
The \textbf{Eliminate} module ($\mathcal{M}_{\textbf{E}}$) uses a zero-shot QA language model to score and mask objects and receptacles that are irrelevant to the current sub-task.
The \textbf{Track} module ($\mathcal{M}_{\textbf{T}}$) uses a zero-shot QA language model to determine if the current sub-task is complete and moves to the next sub-task.
Note that Plan is a generative task and Eliminate and Track are classification tasks.

We also implement an attention-based \textbf{agent} (Action Attention), which scores each permissible action and is trained on imitation learning on the expert. The agent observes the masked observation and takes an action conditioned on the current sub-task. 

\paragraph{Problem Setting} We define the task description as $\mathcal{T}$, the observation string at time step $t$ as $\mathcal{O}^t$, and the list of permissible actions $\left\{a_i^t|a_i^t\text{ can be executed}\right\}$ as $A^t$.
For each observation string $\mathcal{O}^t$, we define the receptacles and objects within the observation as $r_i^t$ and $o_i^t$ respectively. The classification between receptacles and objects is defined by the environment \cite{alfworld}. For a task $\mathcal{T}$, we assume there exists a list of sub-tasks $\mathcal{S}_{\mathcal{T}}=\{s_1, \ldots s_k\}$ that solves $\mathcal{T}$.

\subsection{Plan}

\label{PET:plan}
\begin{figure}[t]
    \centering
    \vspace{-2mm}
    \includegraphics[trim={0cm 0cm 0cm 0cm},width=0.5\textwidth]{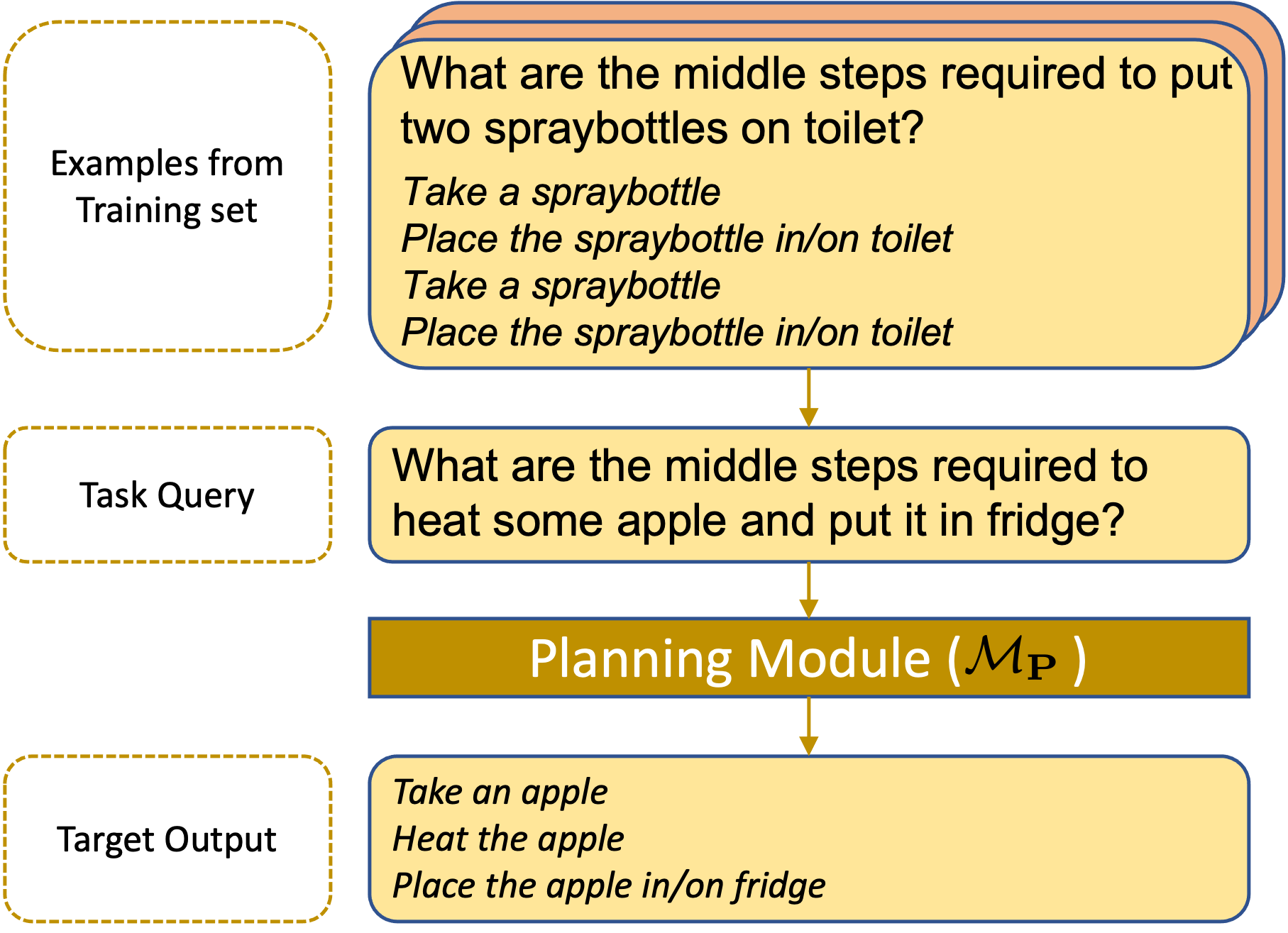} 
    \vspace{-6mm}
    \caption{\small Plan Module (Sub-task Generation). 5 full examples are chosen from the training set based on RoBERTa embedding similarity with the task query description. Then the examples are concatenated with the task query to get the prompt. Finally, we prompt the LLM to generate the desired sub-tasks.}
    \vspace{-3mm}
    \label{fig:plan}
\end{figure}

Tasks in the real world are often complex and need more than one step to be completed. 
% For example the task shown in Fig~\ref{fig:plan}, to accomplish a seemingly simple task such as `heat some apple and put it in the fridge', the agent has to follow 
Motivated by the ability of humans to plan high-level sub-tasks given a complex task, we design the \textbf{Plan} module ($\mathcal{M}_{\textbf{P}}$) to generate a list of high-level sub-tasks for a task description $\mathcal{T}$. %Formally, we define Plan module $\mathcal{P}$ which takes a task description $\mathcal{T}$ as input and generates a list of sub-tasks $\mathcal{S}=\{s_1, \ldots s_k\}$. 

Inspired by the contextual prompting techniques for planning with LLMs \cite{deepak}, we use an LLM as our plan module $\mathcal{M}_{\textbf{P}}$.
For a given task description $\mathcal{T}$, we compose the query question $\mathcal{Q}_{\mathcal{T}}$ as ``What are the middle steps required to $\mathcal{T}$?", and require $\mathcal{M}_{\textbf{P}}$ to generate a list sub-tasks $\mathcal{S}_{\mathcal{T}}=\{s_1, \ldots s_k\}$.

Specifically, we select the top $5$ example tasks $\mathcal{T}^E$ from the training set based on RoBERTa \cite{roberta} embedding similarity with the query task $\mathcal{T}$. We then concatenate the example tasks with example sub-tasks in a query-answer format to build the prompt $\mathcal{P}_{\mathcal{T}}$ for $\mathcal{M}_{\textbf{P}}$ (Fig.~\ref{fig:plan}): 
\[\mathcal{P}_{\mathcal{T}} = \mathtt{concat}(\mathcal{Q}_{\mathcal{T}^E_1}, \mathcal{S}_{\mathcal{T}^E_1}, \ldots, \mathcal{Q}_{\mathcal{T}^E_5}, \mathcal{S}_{\mathcal{T}^E_5}, \mathcal{Q}_{\mathcal{T}})\]

An illustration of our prompt format is shown in Figure~\ref{fig:plan}, where $\mathcal{T}=$``heat some apple and put it in fridge'', and $\mathcal{Q}_{\mathcal{T}^E_1}=$``What are the middle steps required to put two spraybottles on toilet'', $\mathcal{S}_{\mathcal{T}^E_1}=$``take a spraybottle, place the spraybottle in/on toilet, take a spraybottle, place the spraybottle in/on toilet''.
The expected list of sub-tasks to achieve this task $\mathcal{T}$ is $s_1=$`take an apple', $s_2=$`heat the apple', and $s_3=$`place the apple in/on fridge'

% We also augment the input with $k$ examples of tasks that are divided into sub-tasks to teach it (again write it mathematically). The examples are selected based on cosine similarity of BERT embedding between the examples and the evaluation task query description (again write this in equation).

% Previous works have shown that large language models (LLMs) pre-trained on rich data could be queried for a variety of tasks without training \cite{gpt3}. Contextual information is provided as part of the input prompt and the LLM is then tasked to complete the remaining text.

\begin{figure}[ht]
    \centering
    \vspace{-2mm}
    \includegraphics[width=0.5\textwidth]{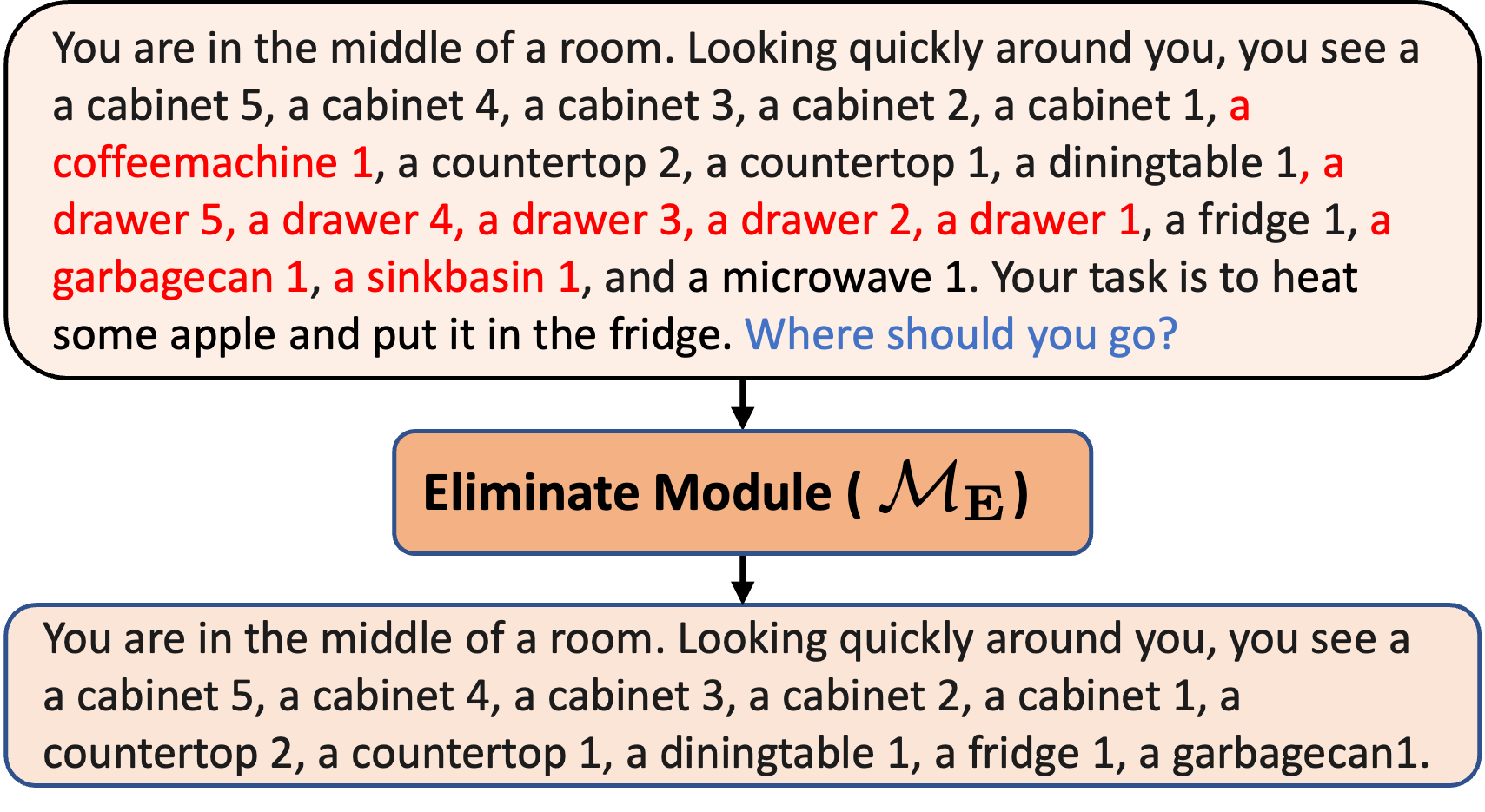} 
    \vspace{-6mm}
    \caption{\small Eliminate Module (Receptacle Masking). We use a pre-trained QA model to filter irrelevant receptacles/objects in the observation of each scene. As we can see, the original observation is too long and the receptacles shown in \textcolor{red}{red} are not relevant for task completion. These receptacles are filtered by the QA model making the observation shorter.}
    \vspace{-4mm}
    \label{fig:eliminate}
\end{figure}

\subsection{Eliminate}
\label{PET:eliminate}
Typical Alfworld scenes can start with around 15 receptacles, each containing up to 15 objects. In some close-to-worst cases, there can be around 30 open-able receptacles (e.g. a kitchen with many cabinets and drawers), and it easily takes an agent with no prior knowledge more than 50 steps for the agent to find the desired object (repeating the process of visiting each receptacle, opening it, closing it).
% An agent starting with no knowledge about where to look for objects that are relevant to solving the task at hand can easily get stuck. 
We observe that many receptacles and objects are irrelevant to specific tasks during both training and evaluation, and can be easily filtered with common-sense knowledge about the tasks.
For example, in Fig.~\ref{fig:eliminate} the task is to heat some apple. By removing the irrelevant receptacles like the coffeemachine, garbagecan, or objects like knife, we could significantly shorten our observation. We therefore propose to leverage commonsense knowledge captured by large pre-trained QA models to design our Eliminate module $\mathcal{M}_{\textbf{E}}$ to mask out irrelevant receptacles and objects.
% Note that we do not fine-tune the pre-trained QA model for our particular task but we use it in a zero-shot manner.
% \TODO{you can write a mathematical formulation of this but atleast make it clear what is the input and expected output. Something like: the elimate module $\mathcal{E}$ consists of a QA model takes the task description $t_i$ and the receptacles $r$ as input and returns a score $s_{r_{i}}$ for each receptacle signifying if the it is relevant to visit for task completion.}

For task $\mathcal{T}$, we create prompts in the format $\mathcal{P}_r=$``Your task is to: $\mathcal{T}$. Where should you go to?" for receptacles and $\mathcal{P}_o=$``Your task is to: $\mathcal{T}$. Which objects will be relevant?" for objects. Using the pre-trained QA model $\mathcal{M}_{\textbf{E}}$ in a zero-shot manner, we compute score $\mu_{o_i} = \mathcal{M}_{\textbf{E}}(\mathcal{P}_o, o_i)$ for each object $o_i$ and $\mu_{r_i} = \mathcal{M}_{\textbf{E}}(\mathcal{P}_o, r_i)$ for each receptacle $r_j$ in observation at every step. $\mu$ represents the belief score of whether the common-sense QA model believes the object/receptacle is relevant to $\mathcal{T}$. We then remove $o_i$ from observation if $\mu_{o_i}<\tau_{o}$, and remove $r_i$ if $\mu_{r_i}<\tau_{r}$. Threshold $\tau_{o}, \tau_{r}$ are hyper-parameters.
% We then obtain scores from the pre-trained QA model $\mathcal{M}$ representing whether the model believe that the receptacle/object is relevant, and we mask out irrelevant receptacles/objects that have scores below a threshold.

\begin{figure}[ht]
    \centering
    \vspace{-2mm}
    \includegraphics[trim={0cm 0cm 0cm 0cm},width=0.5\textwidth]{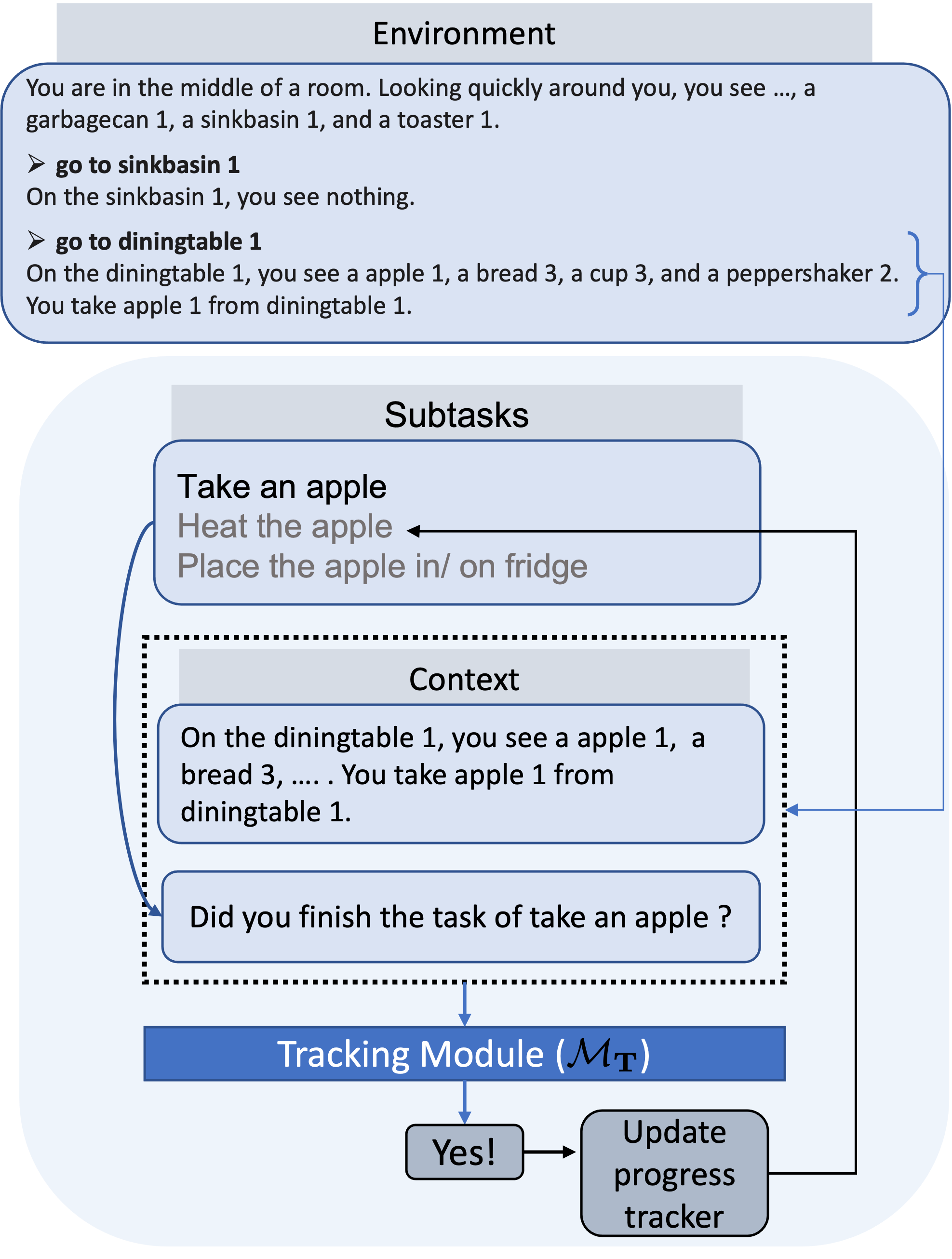} 
    \vspace{-6mm}
    \caption{\small Track Module (Progress Tracking). At every step, we take the last 3 steps of roll-out as context and append a query (about whether the current sub-task is completed) to get the prompt. A pre-trained QA model generates a Yes/No answer to the prompt. For the answer ``Yes", we update the tracker to the next sub-task.}
    \vspace{-4mm}
    \label{fig:track}
\end{figure}

\subsection{Track}
\label{PET:track}
For the agent to utilize the high-level plan, it first needs to know which sub-task to execute. 
% The final challenge for using the high level plan is for the agent to automatically track its progress. 
A human actor typically starts from the first item and check-off the tasks one by one until completion.
% Humans can easily identify the completion of a sub-task using common sense knowledge and observing the trajectory of actions. 
%\amos{The previous sentence has an error (you can write "We observe that..." or "make an observation", buy it still seems awkward). How about something like "Since humans can ..."}
Therefore, similar to Section~\ref{PET:eliminate}, we use a pre-trained QA model to design the Track module $\mathcal{M}_{\textbf{T}}$ to perform zero-shot sub-task completion detection.\footnote{Note that the current system design does not allow re-visiting finished sub-tasks, so the agent has no means to recover if it undoes its previous sub-task at test time.}

Specifically, as illustrated in Figure~\ref{fig:track}, for sub-task list $\mathcal{S}_{\mathcal{T}}=\{s_1, \ldots s_k\}$, we keep track of a progress tracker $p$ (initialized at $1$) that indicates the sub-task the agent is currently working on $(s_p)$. We then compose the context as the last $d$ steps of the agent observation for the current sub-task and the question as ``Did you finish the task of $s_p$?". For efficiency, we set $d := \min(d+1,3)$ at each step. Note that $d$ is reset to $1$ whenever the progress tracker updates. 
Hence, the template $\mathcal{P}_a = \mathtt{concat}(\mathcal{O}^{t-d}, \ldots, \mathcal{O}^{t-1},$ ``Did you finish the task of $s_p$?"$)$.
We feed $\mathcal{P}_a$ to a pre-trained zero-shot QA model $\mathcal{M}_{\textbf{T}}$ and compute the probability of tokens `Yes' and `No' as follows: $p_{\mathcal{M}_{\textbf{T}}}(``Yes"|\mathcal{P}_a)$ and $p_{\mathcal{M}_{\textbf{T}}}(``No"|\mathcal{P}_a)$. 
If $p_{\mathcal{M}_{\textbf{T}}}(``Yes"|\mathcal{P}_a) > p_{\mathcal{M}_{\textbf{T}}}(``No"|\mathcal{P}_a)$ then we increment the tracker $p$ to track the next sub-task. 

If the tracking ends prematurely, meaning that $p>len(\mathcal{S}_{\mathcal{T}})$ but the environment has not returned ``done", we fall back to conditioning with $\mathcal{T}$. We study the rate of pre-mature ends in Section~\ref{track_module_recall} in terms of precision and recall.

%\textcolor{red}{Do we need this sentence?}Note that the current system design does not allow re-visiting finished sub-tasks, so the agent has no means to recover if it somehow undo its previous works at test time. 

\begin{figure*}[t]
    \centering
    \vspace{-2mm}
    \includegraphics[trim={0cm 0cm 0cm 0cm},width=0.9\textwidth]{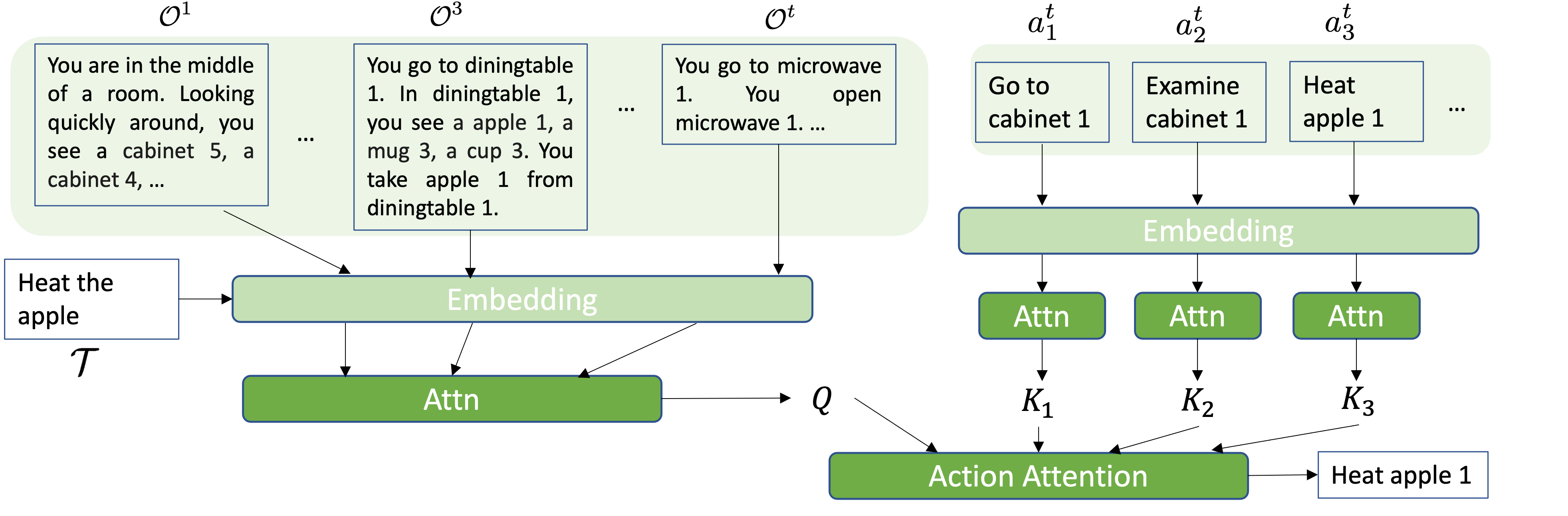} 
    \vspace{-3mm}
    \caption{\small Agent (Action Attention). Action Attention block is a transformer-based framework that computes a key $K_i$ for each permissible action and output action scores as dot-product between key and query $Q$ from the observations.}
    \vspace{-3mm}
    \label{fig:agent}
\end{figure*}

\subsection{Agent}
\label{PET:agent}
Since the number of permissible actions can vary a lot by the environment, the agent needs to handle arbitrary dimensions of action space. While \citet{alfworld} addresses this challenge by generating actions token-by-token, such a generation process leads to degenerate performance even on the training set. 

We draw inspiration from the field of text summarization, where models are built to handle variable input lengths. \citet{pointer} generates a summary through an attention-like ``pointing" mechanism that extracts the output word by word. Similarly, an attention-like ``pointing" model could be used to select an action from the list of permissible actions.

\paragraph{Action Attention} We are interested in learning a policy $\pi$ that outputs the optimal action among permissible actions. We eschew the long rollout/ large action space problems by (1) representing observations by averaging over history, and (2) individually encoding actions (Fig~\ref{fig:agent}).
In our proposed action attention framework, we first represent historical observations $H^t$ as the average of embeddings of all individual observations through history (Eq. 1), and $H^A$ as the list of embeddings of all the current permissible actions (Eq. 2). 
Then, in Eq. 3, we compute the query $Q$ using a transformer with a ``query'' head ($\mathcal{M_{Q}}$) on task embedding ($H^t$), the current observation embedding ($\mathcal{O}^{t}$), and the list of action embeddings ($H^A$). 
In Eq. 4 we compute the key $K_i$ for each action $a_i$ using the same transformer with a ``key'' head ($\mathcal{M_{K}}$) on task embedding ($H^t$), the current observation embedding ($\mathcal{O}^{t}$), and embedding of action ($a_i$).

Finally, we compute the dot-product of the query and keys as action scores for the policy $\pi$ (Eq. 5).
% We propose a transformer-based action attention framework that encodes observation, historical observation, and task string into "query" and calculate attention scores against "key" encoding of permissible actions. Formally, we compute $Q\in \R^d$ using a transformer encoder $\mathcal{M}$ that takes $[Embed(s_{Task}), \text{avg}_{t\in [1,t)} Embed(s^t), Embed(s^T), a_1^T, ..., a_n^T]$ as input and compute $K_i$ with transformer encoder on $[Embed(s_{Task}), \text{avg}_{t\in [1,t)} Embed(s^t), Embed(s^T), a_i^T]$ for all $i$. 
%\TODO{Plain English: }
% \TODO{add more equations to make the flow clear.}
\begin{align}
    H^t &= \text{avg}_{j\in [1,t-1]} \text{Embed}(\mathcal{O}^j)\\
    H^A &= \left[\text{Embed}(a_1^t), ..., \text{Embed}(a_n^t)\right] \\
    Q &= \mathcal{M_{Q}}\left(\text{Embed}(\mathcal{T}), H^t, \text{Embed}(\mathcal{O}^t), H^A\right) \\
    K_i &= \mathcal{M_{K}}\left(\text{Embed}(\mathcal{T}), H^t, \text{Embed}(\mathcal{O}^t), \text{Embed}(a_i^t)\right) \\
    \pi &= \text{softmax}\left([Q \cdot K_i | i \in \text{all permissible actions}]\right)
\end{align}
%according to a decision threshold of $0.4$.

% Para1: Motivation. why there is a need of object masking. hard to find small objects, too many objects that may not be needed for task completion. 

% Para2: Example: Commonsense that to wash a potato you don't need a pan

% Para 3: our solution is to leverage the knowledge that exists in pre-trained models. our approach: what exactly are you asking the model

% Para 4: example of the question-answer 

%% file: sections/results.tex
\section{Experiments and Results}
We present our experiments as follows. First, we explain the environment setup and baselines for our experiments. Then we compare PET to the baselines on different splits of the environment. Finally, we conduct ablation studies and analyze the PET framework part by part. We show that PET generalizes better to human goal specification under efficient behavior cloning training. %\textcolor{red}{Especially, we show that PET is generalizable to human goals from in-simulation training only, unlike existing work. We don't use Dagger... }
% We are solving ... . ALFWorld is a simulator that enables agents to learn abstract, text-based policies in TextWorld \cite{textworld} and then execute goals from the ALFRED benchmark \cite{alfred} in a rich visual environment. \TODO{What do we do about this?}

\subsection{Experimental Details}

\begin{table*}[ht]
\vspace{-2mm}
\begin{center}
\begin{tabular}{l @{\hskip 0.50in} c c @{\hskip 0.30in} c c}
 & \multicolumn{2}{c}{Template Goal Specification} & \multicolumn{2}{c}{Human Goal Specification} \\ 
\toprule
\multicolumn{1}{c}{Model} & \multicolumn{1}{c}{seen} & \multicolumn{1}{c}{unseen} & \multicolumn{1}{c}{seen} & \multicolumn{1}{c}{unseen} \\
\toprule
BUTLER + DAgger* \cite{alfworld} & 40 & 35 & 8 & 3 \\
BUTLER + BC \cite{alfworld} & 10 & 9 & - & - \\ 
GPT \cite{fewshot} & \textbf{91} & \textbf{95} & 42 & 57 \\
PET + Action Attention (Ours) & 70 & 67.5 & \textbf{52.5} & \textbf{60}\\
\bottomrule
\end{tabular}
\end{center}
\vspace{-3mm}
\caption{\label{table:human} Comparison of different models in terms of completion rate per evaluation split (seen and unseen), with and without human annotated goals. PET under-performs GPT on Template goal specifications but generalizes better to human goal specifications. \textbf{*} We include the performance of BUTLER with DAgger for completeness. All other rows are trained without interaction with the environment, MLE for GPT and behavior cloning for BUTLER+BC and PET. }
\end{table*}

\paragraph{AlfWorld Environment}
ALFWorld \cite{alfworld} is a set of TextWorld environments \cite{textworld} that are parallels of the ALFRED embodied dataset \cite{alfred}. ALFWorld includes 6 task types that each require solving multiple compositional sub-goals. There are 3553 training task instances (\{tasktype, object, receptacle, room\}), 140 in-distribution evaluation task instances (seen split - {tasks themselves are novel but take place in rooms seen during training}) and 134 out-of-distribution evaluation task instances (unseen split - {tasks take place in novels rooms}). An example of the task could be: ``Rinse the egg to put it in the microwave." Each training instance in AlfWorld comes with an expert, from which we collected our training demonstration.

\paragraph{Human Goal Specification} The crowd-sourced human goal specifications for evaluation contain 66 unseen verbs and 189 unseen nouns \cite{alfworld}. In comparison, the template goals use only 12 ways of goal specification. In addition, the sentence structure for human goal specification is more diverse compared to the template goals. Therefore, human goal experiments are good for testing the generalization of models to out-of-distribution scenarios.

\paragraph{Pre-trained LMs.} 
For the \textbf{Plan} module (sub-task generation), we experimented with the open-source GPT-Neo-2.7B \cite{gpt-neo}, and an industry-scale LLM with 530B parameters \cite{megatron-530b}.
For the \textbf{Eliminate} module (receptacle/object masking), we choose Macaw-11b \cite{tafjord2021general}, which is reported to have common sense QA performance on par with GPT3 \cite{gpt3} while being orders of magnitudes smaller. We use a decision threshold of $0.4$ for Macaw score below which the objects are masked out. For the \textbf{Track} module (progress tracking), we use the same Macaw-11b model as the Eliminate module answer to Yes/No questions.

\paragraph{Actor Model Design.} 
Our \textbf{Action Attention} agent ($\mathcal{M_{Q}}$ and $\mathcal{M_{K}}$) is a 12-layer transformer with 12 heads and hidden dimension 384. The last layer is then fed into two linear heads to generate $K$ and $Q$. For embedding of actions and observations, we use pre-trained RoBERTa-large \cite{roberta} with embedding dimension 1024. For sub-task generation, we use ground-truth sub-tasks for training, and generated sub-tasks from Plan module for evaluation.
% \paragraph{AlfWorld Environment.} ALFWorld is a simulator that enables agents to learn abstract, text-based policies in TextWorld \cite{textworld} and then execute goals from the ALFRED benchmark \cite{alfred} in a rich visual environment.

\paragraph{Experimental Setup.} \label{experimental_setting} Unlike the original benchmark \cite{alfworld}, we experiment with models trained with behavior cloning.
Although \citet{alfworld} observe that models benefit greatly from DAgger training, DAgger assumes an expert that is well-defined at all possible states, which is inefficient and impractical. In our experiments, training is 100x slower with DAgger compared to behavior cloning (3 weeks for DAgger v.s. 6 hours for Behavior Cloning). 
In addition, we demonstrate that our models surpass the DAgger training performance of the BUTLER \cite{alfworld} agents trained with DAgger, even when our agent does not have the option to interact with the environment.

\paragraph{Baselines.} Our first baseline is the BUTLER::BRAIN (\textbf{BUTLER}) agent \cite{alfworld}, which consists of an encoder, an aggregator, and a decoder. At each time step $t$, the encoder takes initial observation $s^0$, current observation $s^t$, and task string $s_{\text{task}}$ and generates representation $r^t$. The recurrent aggregator combines $r^t$ with the last recurrent state $h^{t-1}$ to produce $h^t$, which is then decoded into a string $a^t$ representing action. In addition, the BUTLER agent uses beam search to get out of stuck conditions in the event of a failed action. Our second baseline \textbf{GPT} \cite{fewshot} is a fine-tuned GPT2-medium on 3553 demonstrations from the AlfWorld training set. Specifically, the GPT is fined-tuned to generate each action step word-by-word to mimic the rule-based expert using the standard maximum likelihood loss.

\subsection{Overall Results on Template and Human Goals}
We compare the performance of action attention assisted by PET with BUTLER \cite{alfworld} and fine-tuned GPT \cite{fewshot} in Table~\ref{table:human}. For human goal specifications, PET outperforms SOTA (GPT) by 25\% on seen and 5\% on the unseen split. 

Although PET under-performs GPT on Template goal specifications, GPT requires fine-tuning on fully text-based expert trajectory and thus loses adaptability to different environment settings. Qualitatively, on human goal specification tasks, where the goal specifications are out-of-distribution, GPT often gets stuck repeating the same action after producing a single wrong move. On the other hand, since the Plan module of PET is not trained on the task, it generalizes to the variations for human goal specifications as shown in Section~\ref{plan_generalization}. Quantitatively, GPT suffers from a relative $50\%$ performance drop transferring from template to human-goal specifications, whereas PET incurs only a $15\sim25\%$ drop.

The setting closest to PET is BUTLER with behavior cloning (BUTLER + BC). Since BUTLER + BC performs poorly, we also include DAgger training results. Nevertheless, action attention assisted by PET outperforms BUTLER with DAgger by more than 2x while being much more efficient. (Section~\ref{experimental_setting})
% Although PET under-performs GPT on Template goal specifications, PET generalize better to human goal specifications\textcolor{red}{should explain what this is, I think} since the Plan module is not trained to fit specific distributions. We observe that PET incurs only a $15\sim25\%$ relative performance drop compared to a $50\%$ relative drop for GPT. We note that a drop in performance cannot be avoided for human goals since some of the human annotations for AlfWorld are incorrect, making some tasks unsolvable.

% \begin{enumerate}
% \item 1: we compare with GPT and we are better at human goals
% \item 2: GPT is not generalizable
% \item 3: Give example of Human goal specification. Explain why GPT fails.
% \item If GPT goes under wrong track it just repeats and does not recover.
% \item separate para: the architecture/paradigm closest to us is BUTLER + BC and we are X\% better than that. Infact, DAgger using a lot more data, we still perform Y\% better than DAgger.
% \end{enumerate}
\begin{table*}[ht]
\vspace{-2mm}
\begin{center}
\begin{tabular}{l @{\hskip 0.30in} c c @{\hskip 0.30in} c c}
 & \multicolumn{2}{c}{Template Goals} & \multicolumn{2}{c}{Human Goals} \\
\toprule
\multicolumn{1}{c}{LLM} & \multicolumn{1}{c}{seen} & \multicolumn{1}{c}{unseen} & \multicolumn{1}{c}{seen} & \multicolumn{1}{c}{unseen} \\
\toprule
GPT-2 \cite{radford2019language} & 94.29 (0.97) & 87.31 (0.94) & 10.07 (0.62) & 7.98 (0.58)\\
GPT-Neo-2.7B \cite{gpt-neo} & \textbf{99.29 (1.00)} & 96.27 (0.98) & 4.70 (0.82) & 9.16 (0.80)\\
MT-NLG \cite{megatron-530b} & 98.57 (0.99) & \textbf{100 (1.00)} & \textbf{40.04 (0.94)} & \textbf{49.3 (0.94)}\\
\bottomrule
\end{tabular}
\end{center}
\vspace{-3mm}
\caption{\label{table:LLMs-plan} Evaluation of different LLMs for \textbf{Plan} module in terms of accuracy and RoBERTa embedding cosine similarity (in brackets) against ground-truth sub-tasks, per evaluation split (seen and unseen), with and without human annotated goals. The MT-NLG with 530B parameters achieves the overall best performance on all dataset splits and greatly exceeds the performance of smaller models on hard tasks with human goal specification. In addition, MT-NLG generates sub-tasks with almost perfect embedding similarity for all tasks.}
\vspace{-2mm}
\end{table*}
\begin{table}[h]
\begin{center}
\begin{tabular}{l @{\hskip 0.30in} c c}
\multicolumn{1}{c}{Model Ablations} & \multicolumn{1}{c}{seen} & \multicolumn{1}{c}{unseen} \\
\toprule
Action Attention & 25 & 9\\
Action Attention + Eliminate & 25 & 11\\
Action Attention + Plan \& Track & 35 & 15\\
Action Attention + PET  & 52.5 & 27.5\\
\bottomrule
\end{tabular}
\end{center}
\vspace{-2mm}
\caption{\label{table:ablations} Comparison of different ablations of PET trained on a sampled set of 140 demonstrations from the training set, in terms of completion rate per evaluation split (seen and unseen). Applying Eliminate module alone has an insignificant effect on overall performance compared to Plan \& Track. However, applying Eliminate module on sub-tasks together with Plan \& Track results in a much more significant performance improvement.}
\vspace{-2mm}
\end{table}

\subsection{Ablations for Plan, Eliminate, and Track}
In Table~\ref{table:ablations}, we analyze the contribution of each PET module by sequentially adding each component to the action attention agent on 140 training trajectories sampled from the training set. The data set size is chosen to match the size of the seen validation set, for an efficient and sparse setting. Note that we treat Plan and Track as a single module for this ablation since they cannot work separately. 

Adding Plan and Track greatly improves the completion rate relatively by $60\%$, which provides evidence to our hypothesis that solving some embodied tasks step-by-step reduces the complexity.
We observe a relatively insignificant $3\%$ improvement in absolute performance when adding Eliminate without sub-task tracking. On the other hand, when applying Eliminate to sub-tasks with Plan and Track, we observe more than $60\%$ relative improvement over just Plan and Track alone. We, therefore, deduce that Plan and Track boost the performance of Eliminate during evaluation, since it is easier to remove irrelevant objects when the objective is more focused on sub-tasks. 

\subsection{Automated Analysis of PET modules}

\paragraph{Plan Module}
We experiment with different LLMs such as GPT2-XL~\cite{radford2019language}, GPT-Neo-2.7B~\cite{gpt-neo}, and the 530B parameter MT-NLG~\cite{megatron-530b} models. 
Table~\ref{table:LLMs-plan} reports the generation accuracy and the RoBERTa \cite{roberta} embedding cosine similarity against ground-truth sub-tasks. We observe that all LLMs achieve high accuracy on template goal specifications, where there is no variation in sentence structures. 
For human goal specification, MT-NLG generates subtasks similar to ground truth in terms of embedding similarity, while the other smaller models perform significantly worse. %We therefore choose the MT-NLG as the Plan module.

\begin{figure*}[ht]
    \centering
    \vspace{-2mm}
    \includegraphics[trim={0cm 0cm 0cm 0cm}, clip, width=0.3\textwidth]{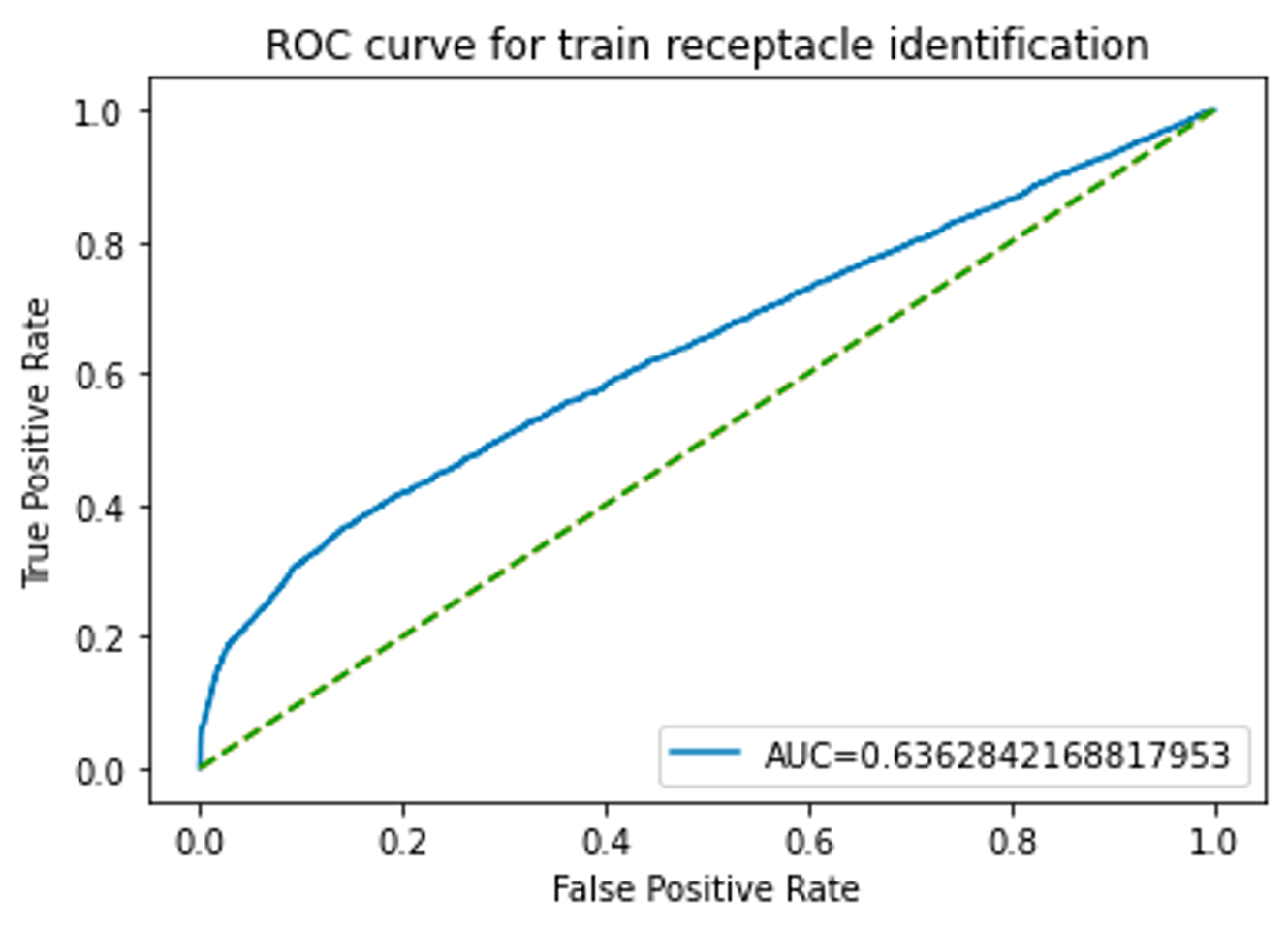}
    \includegraphics[trim={0cm 0cm 0cm 0cm}, clip, width=0.3\textwidth]{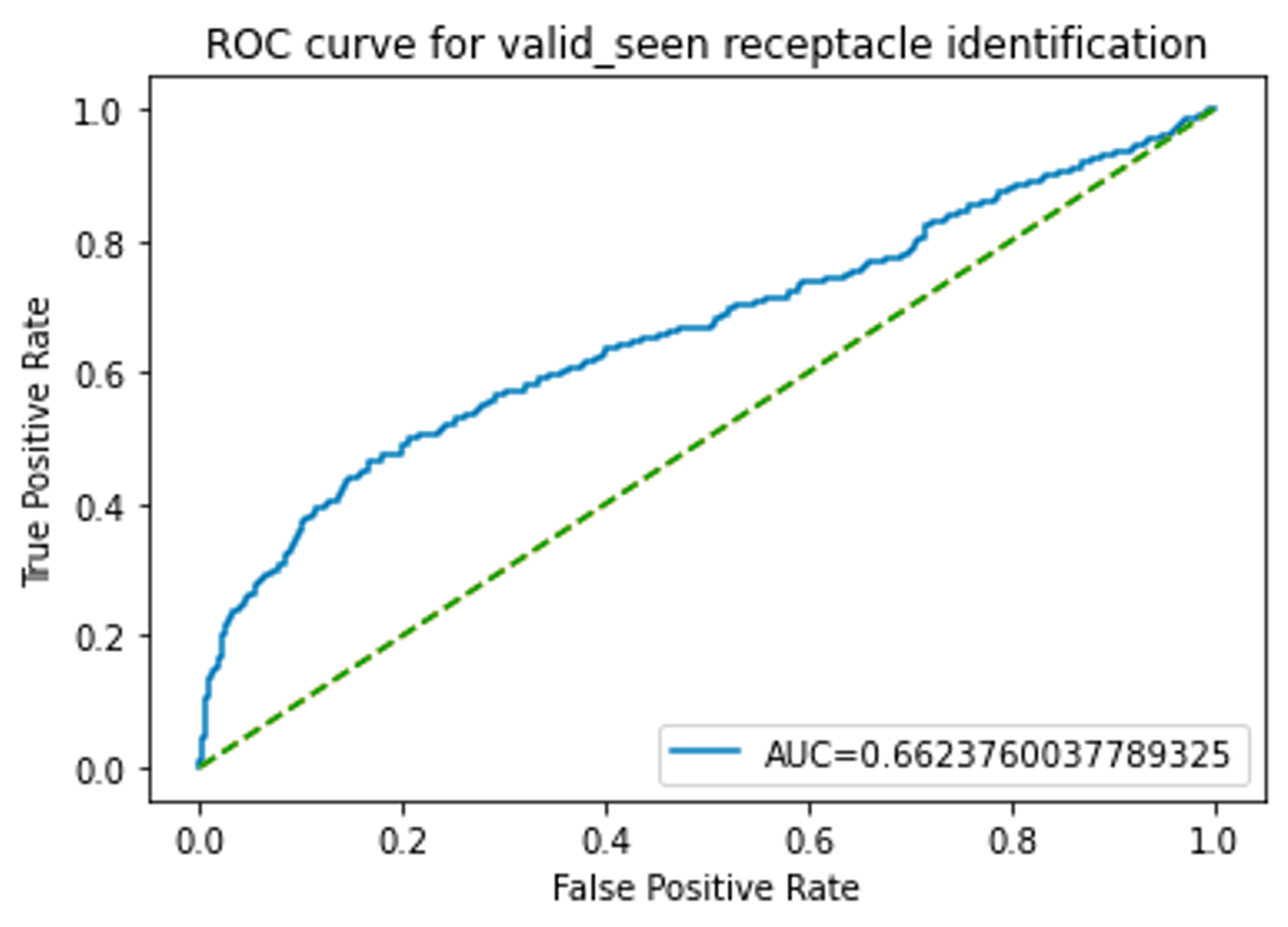}
    \includegraphics[trim={0cm 0cm 0cm 0cm}, clip, width=0.3\textwidth]{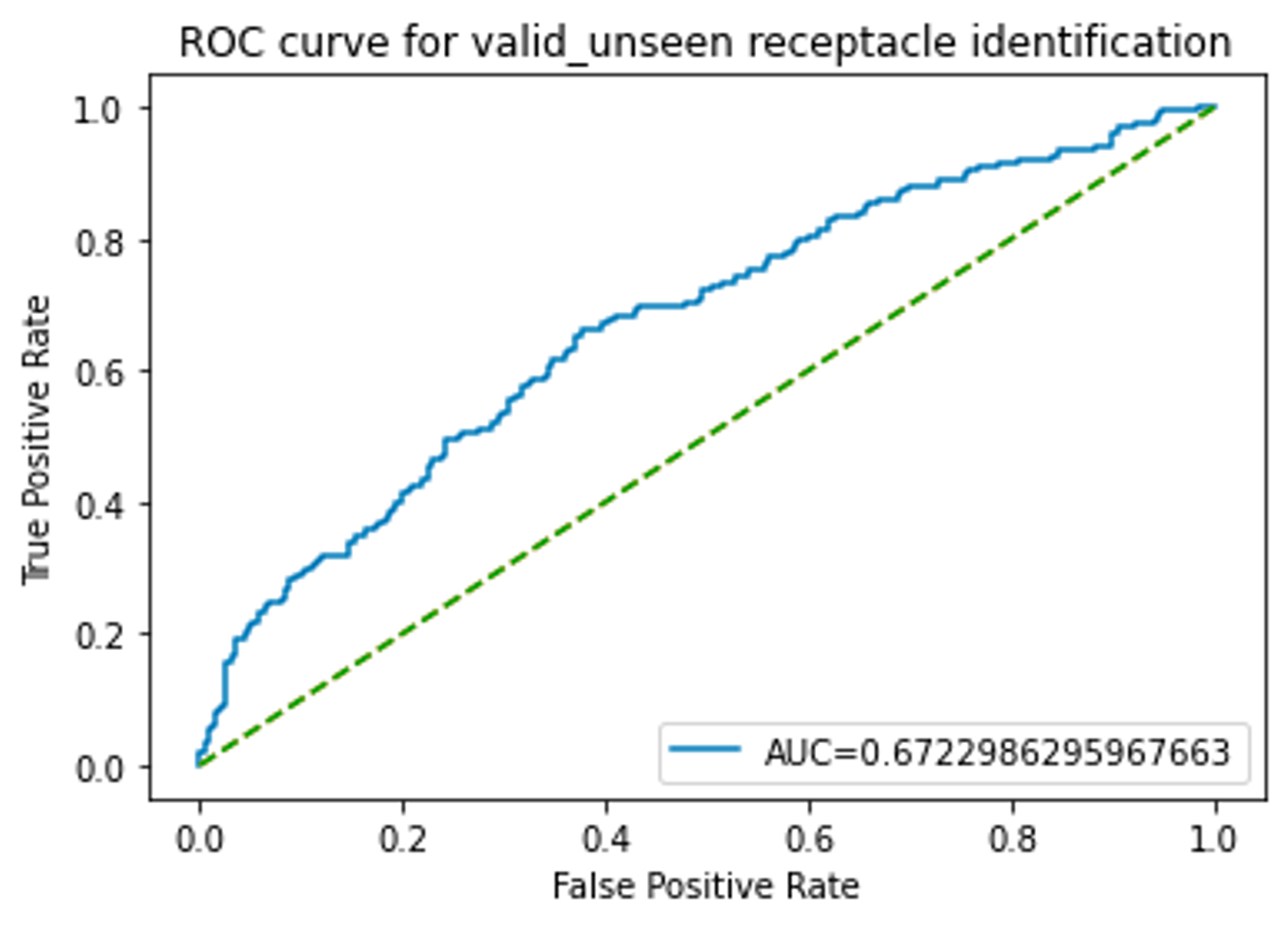}
    \includegraphics[trim={0cm 0cm 0cm 0cm}, clip, width=0.3\textwidth]{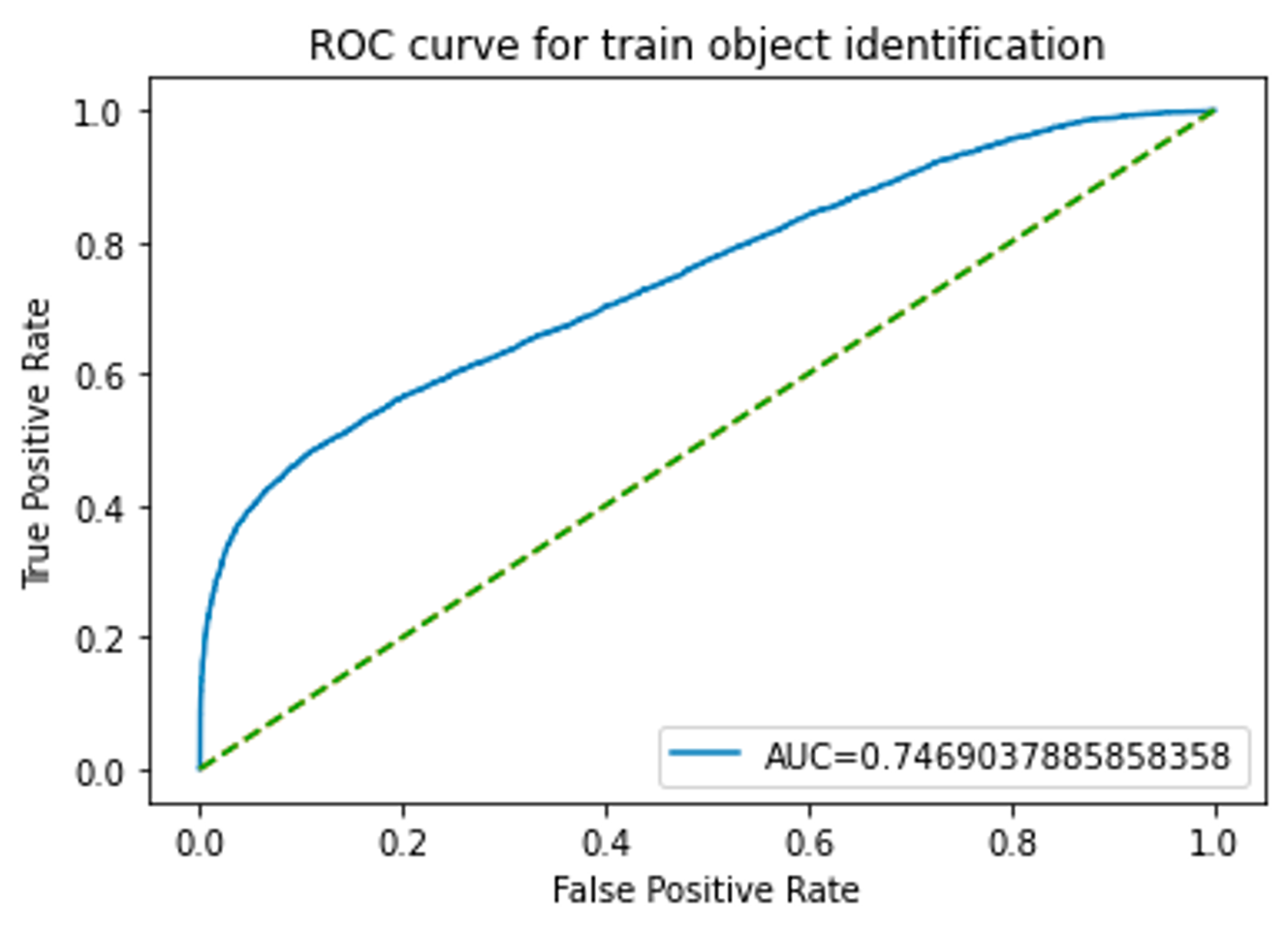} 
    \includegraphics[trim={0cm 0cm 0cm 0cm}, clip, width=0.3\textwidth]{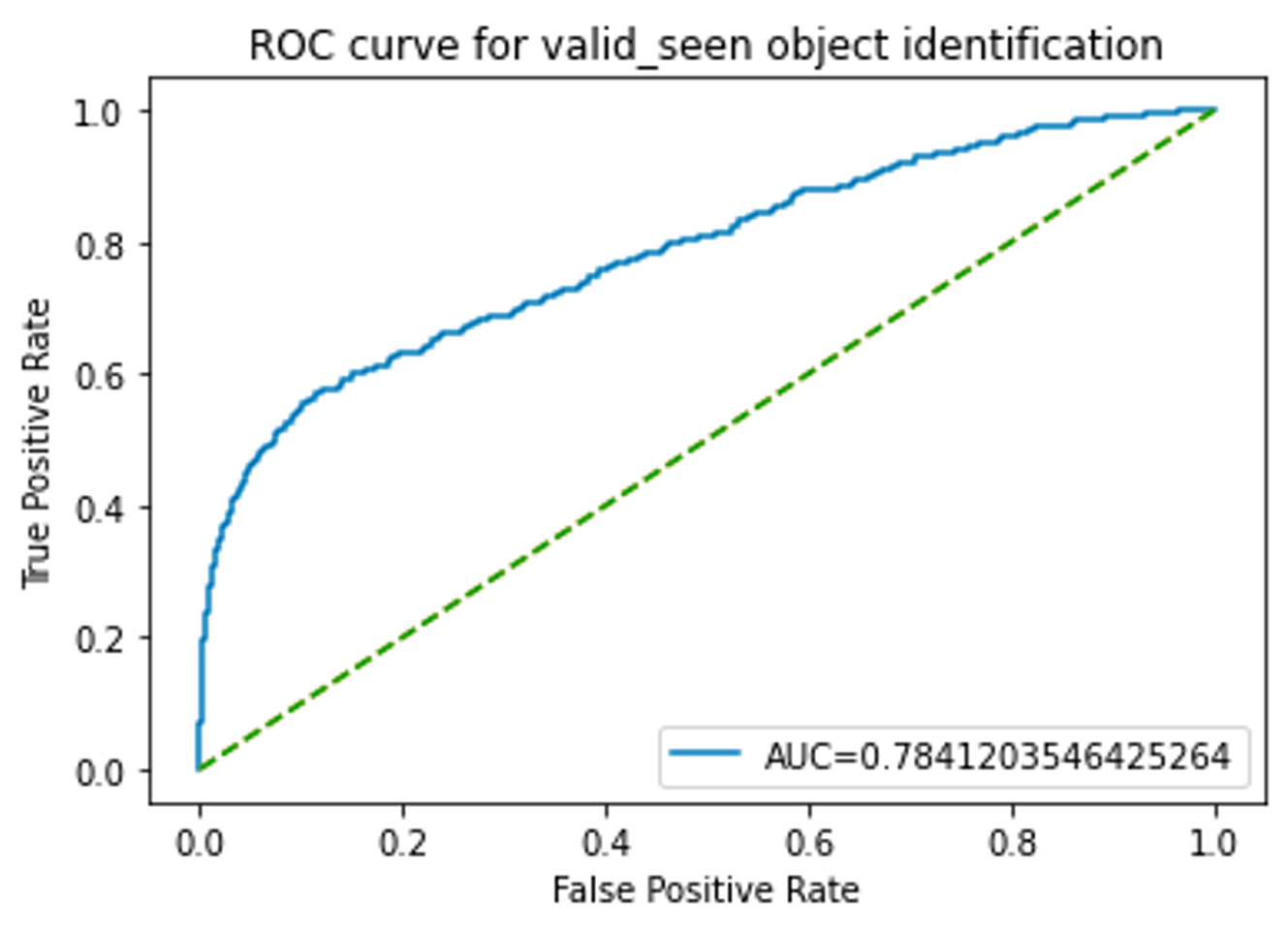}
    \includegraphics[trim={0cm 0cm 0cm 0cm}, clip, width=0.3\textwidth]{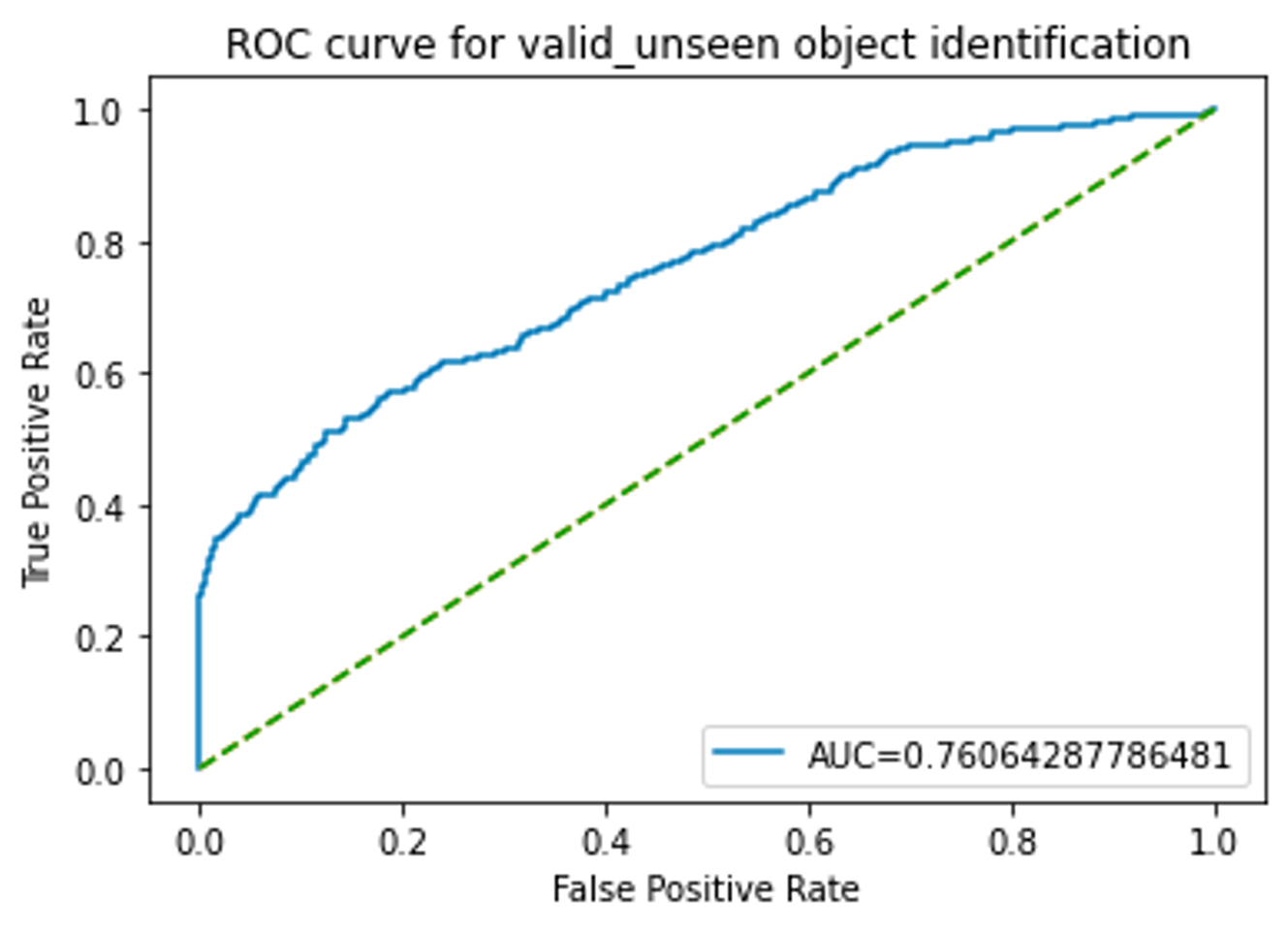}
    \vspace{-4mm}
    \caption{\small Plot of AUC scores of zero-shot relevance identification across all tasks in the Alfworld-Thor environment, with the Macaw-11b model. The ground truth is obtained as receptacles/objects accessed by the rule-based expert. \textbf{Top:} Receptacle relevance identification. \textbf{Bottom:} Object relevance identification. The QA model achieves an average AUC-ROC score of $65$ for receptacles and $76$ on objects.}
    \vspace{-4mm}
    \label{fig:QA_AUC}
\end{figure*}

\paragraph{Eliminate module} 
We evaluate the zero-shot receptacle/object masking performance of Macaw on the three splits of AlfWorld. 
In Fig~\ref{fig:QA_AUC}, we illustrate the AUC curve of the relevance score that the model assigns to the objects v.s. objects that the rule-based expert interacted with when completing each task. Since the Macaw QA model is queried in a zero-shot manner, it demonstrates consistent masking performance on all three splits of the environment, even on the unseen split. In addition, we note that object receptacle accuracy is generally lower than object accuracy because of the counter-intuitive spawning locations described in Section~\ref{eliminate_failure}.
In our experiments, a decision threshold of $0.4$ has a recall of $0.91$ and reduces the number of objects in observation by $40\%$ on average. 
% \TODO{Talk about numbers}
% \TODO{Compare seen/unseen}
% \TODO[Yue]{May need to regenerate graphs}
\vspace{-1mm}
\label{subsec:training}

\paragraph{Track module}
\label{track_module_recall}
Since sub-task alignment information is not provided by the environment, we explore an alternative performance metric for the detection of the event of completion. Ideally, a sub-task tracker should record the last sub-task as ``finished" if and only if the environment is ``fully solved" by the expert. As an agreement measure, we report a precision of 0.99 and a recall of 0.78 for Macaw-11B and a precision of 0.96 and a recall of 0.96 for Macaw-large. The larger model (Macaw-11b) is more precise but misses more detection, therefore limiting the theoretical performance to 78\%. The smaller model is much less accurate according to human evaluation but does not limit the overall model performance in theory. In our experiments, we find that both models produce similar overall results, which may suggest that the overall results could be improved with LLMs doing better on both precision and recall.

%% file: sections/analysis.tex
\begin{table}[h]
{\centering
\begin{tabular}{l m{20em} }
\multicolumn{2}{c}{Human Goal Specification Examples}\\
\toprule
Task & Chill a cup and place it in the cabinet.\\
GT  &\textit{cool} the mug$\rightarrow$\textit{place} the mug \textit{in/on} coffeemachine \\
Gen &\textit{chill} the mug$\rightarrow$\textit{return} the mug \textit{to} coffeemachine \\
\midrule
Task & Take the pencil from the desk, put it on the other side of the desk\\
GT &take a pencil$\rightarrow$place the pencil in/on shelf\\
Gen & pick up the white pencil on the desk$\rightarrow$put the white pencil on another spot on the desk
\end{tabular}
}
\vspace{-2mm}
\caption{\label{table:plan_failures} Failure examples from the Plan module on human goal specifications (Task), ground-truth (GT) v.s. generated (Gen). In the first example, generated plan differs from the ground truth but the meaning agrees. In the second example, the generated plan largely differs from the ground truth due to the mistake in human goal specification --- ``another side on the desk" instead of ``shelf".}
\vspace{-2mm}
\end{table}

\subsection{Qualitative Analysis}

\paragraph{Plan Module}
\label{plan_generalization}
We show two types of failure examples for sub-task generation in Table~\ref{table:plan_failures}. The first type of error is caused by generating synonyms of the ground truth, and the second type of error is caused by inaccuracies in the human goal specifications. Note that our Action Attention framework uses RoBERTa \cite{roberta} embedding for sub-tasks, known to be robust to synonym variations.

\paragraph{Eliminate Module}
\label{eliminate_failure}
We observe that the main source of elimination error occurs when the module incorrectly masks a receptacle that contains the object of interest so the agent fails to find such receptacles. This is often because some objects in the AI2Thor simulator do not spawn according to common sense. As noted in the documentation of the environment\footnote{\href{https://ai2thor.allenai.org/ithor/documentation/objects/object-types/}{ai2thor.allenai.org/ithor/documentation/objects/object-types/}}, objects like Apple or Egg has a chance of spawning in unexpected receptacles like GarbageCan, or TVStand. However, such generations in AI2Thor are unlikely in real deployment; thus, the ``mistakes'' of our Eliminate module are reasonable.

\paragraph{Track Module}
Experimentally, we find that sub-task planning/tracking is particularly helpful for tasks that require counting procedures. As shown in Table~\ref{table:rollout}, PET breaks the task of ``Place two soapbar in cabinet" into two repeating set of sub-tasks: ``take soapbar$\rightarrow$place soapbar in/on cabinet". Sub-task planning and tracking, therefore, simplify the hard problem of counting.

%% file: sections/conclusion.tex
\section{Conclusion, Limitations, and Future Work}
In this work, we propose the Plan, Eliminate, and Track (PET) framework that uses pre-trained LLMs to assist an embodied agent in three steps. 
% The Plan module generates a list of sub-tasks as the high-level plan for a given task description. 
% The Eliminate module masks out irrelevant objects and receptacles from the observation of the current sub-task. 
% The Track module determines whether the agent has accomplished the current sub-task and moves to the next sub-task. 
Our PET framework requires no fine-tuning and is designed to be compatible with any goal-conditional embodied agents.

%and demonstrate better performance and good generalization to human goal specifications on the AlfWorld \cite{alfworld} benchmark.
In our experiments, we combine PET with a novel Action Attention agent that handles the dynamic action space in AlfWorld. Our Action Attention agent greatly outperforms the BUTLER baseline. In addition, since the PET framework is not trained to fit the training set tasks, it demonstrates better generalization to unseen human goal specification tasks. Finally, our ablation studies show the Plan and Track modules together improve the performance of Eliminate module to achieve the best performance.

Our results show that LLMs can be a good source of common sense and procedural knowledge for embodied agents, and multiple LLMs may be used in coordination with each other to further improve effectiveness.

One of the major limitations of our current system design is that the Track module (progress tracker) does not re-visit finished sub-tasks. If for example, the agent is executing sub-tasks [picked up a pan, put the pan on countertop], and it picked up a pan but put it in the fridge (undo pickup action). Since the progress tracker does not take into consideration previous progress being undone, the system may break in this situation. 
Future work can focus on adding sub-task-level dynamic re-planning to address this limitation or explore other ways in which LLMs can assist the learning of the policy (i.e., reading an instruction manual about the environment).

%% file: sections/appendix.tex
\section{Example Trajectory}
\begin{table}[h]
{\centering
\begin{tabular}{m{8em} m{37em} }
\multicolumn{1}{c}{Sub-task} & \multicolumn{1}{c}{Observation + Action}\\
\hline
\multirow{12}{8em}{take a soapbar} & Looking quickly around you, you see a cabinet 4, a cabinet 3, a cabinet 2, a cabinet 1, a countertop 1, a garbagecan 1, a handtowelholder 2, a handtowelholder 1, a sinkbasin 2, a sinkbasin 1, a toilet 1, and a towelholder 1.\\
& Action: You go to toilet 1.\\
\cline{2-2}
& On the toilet 1, you see \textbf{nothing}.\\
& Action: You go to sinkbasin 1.\\
\cline{2-2}
& On the sinkbasin 1, you see \textbf{nothing}.\\
& Action: You go to sinkbasin 2.\\
\cline{2-2}
& On the sinkbasin 2, you see \textbf{nothing}.\\
& Action: You go to garbagecan 1.\\
\cline{2-2}
& On the garbagecan 1, you see \textbf{nothing}.\\
& Action: You go to countertop 1.\\
\cline{2-2}
& On the countertop 1, you see a soapbar 2, and a soapbar 1.\\
& Action: You take soapbar 1 from countertop 1.\\
\cline{1-2}
\multirow{6}{8em}{place the soapbar in/on cabinet} & You pick up the soapbar 1 from the countertop 1.\\
& Action: You go to cabinet 1.\\
\cline{2-2}
& The cabinet 1 is closed. You open cabinet 1\\
& Action: You open cabinet 1.\\
\cline{2-2}
& The cabinet 1 is open. In it, you see a cloth 1.\\
& Action: You put soapbar 1 in/on cabinet 1.\\
\cline{1-2}
\multirow{6}{8em}{take a soapbar} & You put the soapbar 1 in/on the cabinet 1.\\
& Action: You close cabinet 1.\\
\cline{2-2}
& You close the cabinet 1.\\
& Action: You go to countertop 1.\\
\cline{2-2}
& On the countertop 1, you see a soapbar 2.\\
& Action: You take soapbar 2 from countertop 1.\\
\cline{1-2}
\multirow{6}{8em}{place the soapbar in/on cabinet} & You pick up the soapbar 2 from the countertop 1\\
& Action: You go to cabinet 1.\\
\cline{2-2}
& The cabinet 1 is closed. You open cabinet 1\\
& Action: You open cabinet 1.\\
\cline{2-2}
& The cabinet 1 is open. In it, you see a cloth 1.\\
& Action: You put soapbar 1 in/on cabinet 1.\\
\cline{1-2}
\end{tabular}
}
\vspace{-10pt}
\caption{\label{table:rollout} Example trajectory of PET agent on the task ``Place two soapbar in cabinet" from unseen evaluation split. The Plan module breaks down the task into two repeating procedures, and the Track module accurately identifies when to switch to the next sub-task, thus avoiding the challenging task of counting. We also observe the Eliminate module at work, removing all irrelevant objects at steps 2,3,4,5, showing \textbf{nothing}.}
\end{table}